\documentclass[runningheads,a4paper]{llncs}
\usepackage[T1]{fontenc}
\usepackage{graphicx}
\usepackage{amsmath}
\usepackage{amssymb}
\usepackage{booktabs}
\usepackage{xcolor}
\usepackage{multirow}
\usepackage{hyperref}

\begin{document}

\title{Beyond Volume Overlap: Surface Matching\\
for Topology-Aware Coronary Artery Segmentation}
\titlerunning{Surface Matching for Coronary Artery Segmentation}


\author{Rafael Velasquez\inst{1}\orcidID{0009-0005-4459-3164} \and 
Esther Puyol-Ant\'{o}n\inst{2}\orcidID{0000-0002-9789-6629} \and
Pablo Arbeláez\inst{1}\orcidID{0000-0001-5244-2407}}

\authorrunning{R. Velasquez et al.}

\institute{
Universidad de los Andes, Bogotá, Colombia
\and
School of Biomedical Engineering \& Imaging Sciences, King's College London, London, UK}

\maketitle
 
\begin{abstract}
 
Accurate coronary artery segmentation on coronary computed tomography angiography (CCTA) is essential for diagnosing coronary artery disease. Deep networks are conventionally trained and evaluated with the Dice coefficient, but volume-overlap metrics are poorly suited to thin, tubular anatomy: since most voxels belong to a few thick proximal segments, a missing distal branch barely affects Dice despite severely disrupting the connectivity required for clinical use. We introduce a surface metric that matches predicted and reference surface points via bipartite assignment under a localized, vessel-radius tolerance, reporting precision, recall, and $F_1$ with decoupled false positives (spurious branches) and false negatives (missed branches)---a distinction the symmetric Dice cannot make. With it we show that a strong Dice-trained baseline omits far more vessel surface than it hallucinates, an asymmetry its high Dice hides. Building on this, we propose a differentiable surface loss that simultaneously suppresses spurious mass and recovers absent structure, validated by fine-tuning three backbones (nnU-Net, SwinUNETR, NexToU) on two public benchmarks (ImageCAS, ASOCA). Against a matched-epoch control, it significantly improves surface $F_1$ by recovering missed distal vessels at comparable Dice. Our findings argue for measuring and optimizing the vessel surface, not the volume it overlaps. Code: \url{https://github.com/BCV-Uniandes/Coronary-Surface-Matching}.
 
\end{abstract}
 
\section{Introduction}
\label{sec:intro}
 
 
Coronary artery disease, driven primarily by arterial stenosis, remains a leading cause of death worldwide~\cite{imagecas2023}. In clinical workflows, Coronary Computed Tomography Angiography (CCTA) provides a non-invasive view of the coronary tree. Highly accurate automated segmentation of this anatomy underpins critical downstream diagnostic pipelines, including automated stenosis grading~\cite{boogers2010} and hemodynamic (FFR-CT) simulation~\cite{ctffr_meta2018}. While deep architectures---ranging from nnU-Net~\cite{nnunet2021} to vision transformers~\cite{swinunetr2022} and topology-aware networks~\cite{nextou2023}---achieve high volumetric scores, they suffer from a fundamental methodological bottleneck: both training and evaluation rely almost exclusively on the volume-overlap Dice coefficient. We argue that this paradigm is fundamentally ill-suited for thin, tubular vascular structures.

Downstream clinical utility, such as CT-derived fractional flow reserve (FFR-CT) modeling, requires complete, topologically connected vessel trees~\cite{gharleghi2022review,ctffr_meta2018}. Centerline-based automated stenosis detection pipelines completely fail when encountering fragmented vessel masks~\cite{boogers2010}. Despite this requirement, the ubiquity of the Dice coefficient introduces three intrinsic failure modes that disproportionately penalize vital distal vessels:
\begin{enumerate}
    \item \textbf{Volume Bias:} Dice is dominated by thick proximal segments, rendering the loss function mathematically blind to thin distal branches that contribute negligibly to total volume.
    \item \textbf{Topological Blindness:} A single broken branch or a hallucinated bridge alters only a handful of voxels, causing negligible changes in Dice while completely corrupting the clinical anatomy~\cite{cldice2021}.
    \item \textbf{Error Conflation:} Dice collapses false positives (spurious branches) and false negatives (missed vessels) into a single symmetric number, failing to diagnose whether a model hallucinates or omits structure.
\end{enumerate}
 
Consequently, a model can yield a deceptively high Dice score while systematically dropping distal vasculature (substantiated in Sec.~\ref{sec:results-dice}). Topology-aware scores partly address this: clDice~\cite{cldice2021} exposes two separable components (topology precision and sensitivity) and topology-aware losses~\cite{topoloss2019} penalise Betti errors. Our metric differs not in having a decomposition but in two respects: its precision and recall are computed over the full vessel surface under a per-point radius tolerance, localising missing or spurious mass at the scale of the local calibre, whereas clDice's components are defined on the skeleton and usually reported through their harmonic mean; and our false positives and negatives are absolute, decoupled point counts rather than a normalised overlap. We use clDice as a strong topology-aware baseline throughout.

Because thin vessels behave as near-pure boundaries, we reformulate coronary segmentation around a surface-to-surface matching paradigm that exposes connectivity errors explicitly: missed branches leave reference surface points unpaired (penalizing recall), hallucinated branches leave predicted points unpaired (penalizing precision). We synthesize contour detection under distance tolerance~\cite{gpb2011} and set prediction via bipartite matching~\cite{detr2020} into a 3D surface framework, replacing fixed tolerances with a \textit{localized, vessel-radius adaptive tolerance} so thin distal branches demand sub-millimeter agreement.
 
\noindent \textbf{Our contributions are twofold:}
\begin{itemize}
    \item \textbf{An Asymmetric Surface Metric (Sec.~\ref{sec:metric}):} A bipartite matching framework between surface points under local vessel-radius constraints that outputs un-subsampled, decoupled precision, recall, and $F_1$ scores to isolate false positives from false negatives.
    \item \textbf{A Differentiable Surface Loss (Sec.~\ref{sec:loss}):} A novel two-term objective function whose gradients explicitly recover missed distal geometry while suppressing spurious artifacts, expanding upon traditional boundary-distance losses~\cite{kervadec2019,cbdice2024}.
\end{itemize}
We validate our method on two public CCTA benchmarks, ImageCAS~\cite{imagecas2023} and ASOCA~\cite{asoca2022}, across three diverse backbones~\cite{nnunet2021,swinunetr2022,nextou2023}, strictly isolating the optimization objective from the network architecture~\cite{cbdice2024}. The source code for the surface metric, surface loss, and experiments reported
in this paper is publicly available at
\url{https://github.com/BCV-Uniandes/Coronary-Surface-Matching}.
\section{Method}
\label{sec:method}
 
\begin{figure}[t]
\centering
\includegraphics[width=0.85\textwidth]{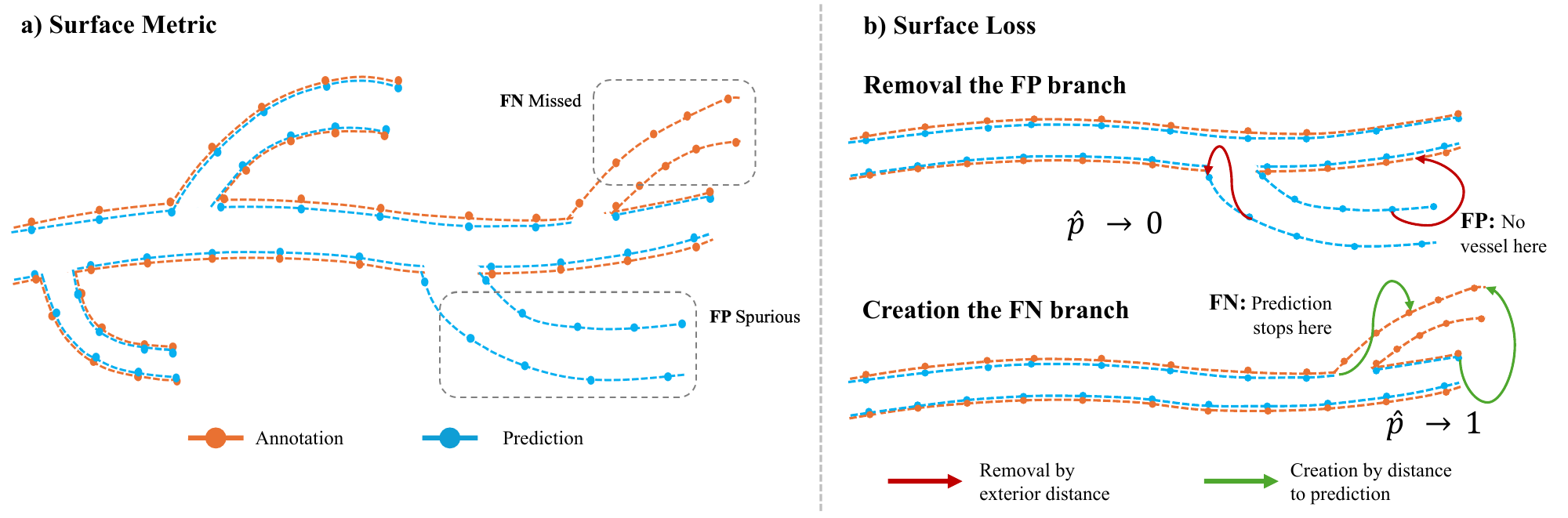}
\caption{Our two contributions, evaluated on identical error modes. \textbf{(a)} The surface metric matches predicted (blue, dashed) and reference (gold) surface points under a per-point radius tolerance $r_i$, decoupling missed branches (FN) from spurious branches (FP) while accepted matches count as true positives (green). \textbf{(b)} The surface loss corrects both: a removal term drives the spurious branch down ($\hat p\!\to\!0$) and a creation term grows the missed branch ($\hat p\!\to\!1$), each weighted by a distance field (Eqs.~\ref{eq:loss-rem},~\ref{eq:loss-cre}). Both terms act directly on surface points, ensuring a thin distal branch receives equal gradient magnitude as a thick proximal trunk.}
\label{fig:method}
\end{figure}

Our model-agnostic framework introduces a bipartite surface metric (Sec.~\ref{sec:metric}) and a differentiable surface loss (Sec.~\ref{sec:loss}) operating directly on raw voxel masks or probability maps. Let $M_{\text{ref}}, M_{\text{pred}} \subset \mathbb{R}^3$ be the reference and predicted volumes. We extract their continuous boundary surfaces, $\mathcal{S}_{\text{ref}} = \{x_i\}_{i=1}^{N}$ and $\mathcal{S}_{\text{pred}} = \{\hat x_j\}_{j=1}^{M}$, via Marching Cubes~\cite{lorensen1987marching}.To capture local anatomical scale, we assign each reference surface point $x_i$ the true local vessel radius $r_i$. Sampling the Euclidean Distance Transform (EDT) of $M_{\text{ref}}$ directly at a surface point would yield a value close to zero, since surface points lie on the vessel boundary; it does not give the radius. Instead, we exploit the fact that the EDT attains its local maximum along the vessel medial axis, where its value equals the distance from the centreline to the wall, i.e.\ the vessel radius. We therefore (i)~compute the EDT of $M_{\text{ref}}$, (ii)~extract the medial ridge as the voxels where the EDT equals its local maximum (a $3\times3\times3$ maximum filter), whose EDT value is the local radius, and (iii)~assign to each surface point $x_i$ the radius of the nearest ridge voxel via a $k$-d tree. This yields, at every surface point, the radius of the vessel it bounds, not a near-boundary value, and requires no centreline annotation. The radius dynamically sets the adaptive matching tolerance for the bipartite metric assignment, as illustrated in Fig.~\ref{fig:method}. Because the ridge is defined on the reference mask alone, $r_i$ is a property of the ground-truth anatomy, independent of any prediction or slice orientation: the EDT and its medial ridge are computed in the volume with physical voxel spacing, so the radius reflects the true 3D vessel calibre regardless of how the vessel is angled relative to the acquisition slices.
 
 
\subsection{Bipartite Surface Metric}
\label{sec:metric}
Let $\{x_i\}_{i=1}^{N}$ denote the reference surface points with local radii $r_i$, and let $\{\hat x_j\}_{j=1}^{M}$ denote the predicted surface points. Crucially, $N$ and $M$ are the \emph{full} point counts of each surface; we do not subsample to a common size, because doing so would force $N=M$ and mechanically equate false positives and false negatives. We accept a pair $(i,j)$ as a true match only when the predicted point falls inside the reference tube, $\lVert x_i-\hat x_j\rVert_2 \le r_i$ (Fig.~\ref{fig:method}a). We then resolve the one-to-one assignment greedily in order of increasing distance, which is near-optimal for this acceptance rule and scales to a full volume's surface points through a $k$-d tree. Unmatched reference points are false negatives (missed branches) and unmatched predicted points are false positives (spurious branches). With $\mathcal{M}$ the set of accepted matches,
\begin{equation}
  \mathrm{Prec}=\frac{|\mathcal{M}|}{M},\qquad
  \mathrm{Rec}=\frac{|\mathcal{M}|}{N},\qquad
  F_1=\frac{2\,\mathrm{Prec}\cdot\mathrm{Rec}}{\mathrm{Prec}+\mathrm{Rec}}.
  \label{eq:metric-prf}
\end{equation}
Because $N$ and $M$ are independent, $\mathrm{FP}=M-|\mathcal{M}|$ and $\mathrm{FN}=N-|\mathcal{M}|$ are decoupled and measure distinct failure modes, and the per-point radius tolerance adapts the metric to vessel calibre. This decoupling lets the metric reveal the hallucinate-versus-omit asymmetry that Dice cannot (Sec.~\ref{sec:results-dice}). Crucially, the tolerance $r_i$ is not a free hyperparameter tuned for our results: it is read directly from the reference distance transform, so at every point it is the true local vessel radius rather than a value we choose. The same anatomically-derived tolerance is applied identically to every method compared, so the metric cannot be tuned toward any particular loss. A systematic sensitivity analysis of the tolerance (e.g.\ scaling $r_i$ between $0.5\times$ and $2\times$) would further characterise the metric's stability and is left to future work.
 
\subsection{Surface Loss}
\label{sec:loss}
A voxel-wise objective scores every voxel independently and never sees \emph{where} the predicted boundary lies. We instead supervise the prediction with two distance-based terms mirroring the two error modes the metric decouples, namely spurious mass to be removed and missed structure to be created (Fig.~\ref{fig:method}b). Let $g\in\{0,1\}^\Omega$ be the reference mask on the voxel grid $\Omega$ and $\hat p:\Omega\to[0,1]$ the predicted vessel probability, with voxel spacing $s$ so that distances are in millimetres. For a non-empty voxel set $A$, let $d_A(v)=s\,\min_{u\in A}\lVert v-u\rVert_2$ be the Euclidean distance field from each voxel $v$ to $A$, in millimetres; both terms below are instances of this field.
 
\subsubsection{Removal term (false positives).}
From $g$ we precompute, once per case and without gradient, the exterior distance field $\phi_{\mathrm{out}}=d_{g}$ evaluated outside the vessel and set to zero inside it ($v\in g$), optionally clipped at $\delta$~mm. The removal term penalises probability placed far from the vessel,
\begin{equation}
  \mathcal{L}_{\mathrm{rem}}(\hat p,g)=
  \frac{1}{N_{\mathrm{bg}}}\sum_{v\in\Omega} \phi_{\mathrm{out}}(v)\,\hat p(v),
  \label{eq:loss-rem}
\end{equation}
where $N_{\mathrm{bg}}$ is the number of background voxels. Its gradient is largest for spurious probability far from the reference surface, driving it down.
 
\subsubsection{Creation term (false negatives).}
To create missed structure we penalise low probability on the reference surface, weighted by how far the prediction currently is from covering it. Let $S\subset\Omega$ be the reference surface voxels and, at each step, let $D_{\hat p}=d_{\{u:\hat p(u)>1/2\}}$ be the distance field to the current predicted vessel mass, recomputed from the prediction and treated as a constant field (no gradient). The creation term is
\begin{equation}
  \mathcal{L}_{\mathrm{cre}}(\hat p,g)=
  \frac{1}{|S|}\sum_{v\in S} D_{\hat p}(v)\,\bigl(1-\hat p(v)\bigr).
  \label{eq:loss-cre}
\end{equation}
Because $D_{\hat p}$ is a fixed field at each step, the term retains leverage even where $\hat p\!\approx\!0$: an entirely-missed branch lies far from the predicted mass, so $D_{\hat p}$ is large there and the term pushes probability up along the whole missed branch. This differs from an interior distance field, whose weight peaks at the deep core of thick vessels and so reproduces the volume-dominance Dice suffers from. Weighting by surface rather than interior volume treats thin distal and thick proximal segments on equal footing.
 
\subsubsection{Total objective.}
We add the combined surface loss to the standard deep-supervised backbone loss~\cite{nnunet2021} with a linear warm-up,
\begin{equation}
  \mathcal{L}_{\mathrm{c}}=
  \lambda_{\mathrm{rem}}\,\mathcal{L}_{\mathrm{rem}}
  +\lambda_{\mathrm{cre}}\,\mathcal{L}_{\mathrm{cre}},
  \qquad
  \mathcal{L}=\mathcal{L}_{\mathrm{Dice+CE}}+\lambda(t)\,\tilde{\mathcal{L}}_{\mathrm{c}},
  \quad
  \lambda(t)=\lambda_{\max}\,\min\!\Bigl(1,\tfrac{t}{T_w}\Bigr),
  \label{eq:objective}
\end{equation}
where $t$ indexes the epoch and $T_w$ the warm-up length. We rescale the surface term by a detached running estimate of its magnitude relative to the base loss, $\tilde{\mathcal{L}}_{\mathrm{c}}=\mathcal{L}_{\mathrm{c}}\cdot \mathcal{L}_{\mathrm{Dice+CE}}/\mathrm{EMA}(\mathcal{L}_{\mathrm{c}})$, so $\lambda_{\max}$ is the target fraction of the base loss and needs no manual tuning. We apply the surface term to the highest-resolution deep-supervision head. Because the loss supervises by dense distance fields while the metric solves a hard matching at evaluation, a metric improvement cannot be an artefact of optimising it.
 
\textbf{Relation to boundary and distance losses.} Our loss belongs to the distance-transform family, of which Kervadec et al.'s boundary loss~\cite{kervadec2019} is canonical, but differs in three ways that matter for thin distal vessels. First, the boundary loss is a \emph{single} signed-distance term and does not separate the two error modes; ours uses \emph{two decoupled} terms, removal (false positives) and creation (false negatives), weighting missed branches independently of spurious ones. Second, and most important for recovery, the boundary loss weights each voxel by its distance to the \emph{reference} boundary, fixed throughout training; our creation term instead uses the distance to the \emph{current} prediction $D_{\hat p}$, recomputed each step, so an entirely-missed branch receives a large, growing gradient until covered rather than a near-constant pull. Third, it integrates the interior distance field, whose magnitude scales with calibre and reintroduces the volume dominance we avoid, whereas our terms act on surface points. The consequence is visible in Table~\ref{tab:imagecas}: against cbDice, the strongest distance-transform baseline, our loss leaves over $5{,}000$ fewer missed surface points.
 
\section{Experiments}
\label{sec:experiments}
 
\subsubsection{Datasets.}
ImageCAS~\cite{imagecas2023} provides 1000 CCTA volumes with binary annotations; we hold out 200 volumes for testing and run 5-fold cross-validation on the remaining 800. ASOCA~\cite{asoca2022} provides 40 annotated CCTA cases (20 normal, 20 diseased); we partition the 40 cases ourselves, holding out 8 (20\%) for testing and running 5-fold cross-validation on the remaining 32, stratified to preserve the normal/diseased balance. Both datasets are publicly available and fully de-identified by their providers under their respective licenses; our use is secondary analysis of public data and required no additional ethics approval.
 
\subsubsection{Backbones, training, and baselines.}
We evaluate three backbones, nnU-Net~\cite{nnunet2021}, SwinUNETR~\cite{swinunetr2022}, and NexToU~\cite{nextou2023}, each in its 3D full-resolution configuration within the nnU-Net framework. For every backbone we train a Dice+CE baseline for $1000$ epochs, then fine-tune from those weights for a further $150$ epochs at learning rate $10^{-3}$. We compare four configurations per backbone: the Dice+CE baseline, clDice~\cite{cldice2021}, cbDice~\cite{cbdice2024}, and our surface loss, all fine-tuned under this common protocol. clDice~\cite{cldice2021} supervises the overlap of the morphological skeletons of prediction and reference, rewarding centreline connectivity; cbDice~\cite{cbdice2024} extends this by weighting the centreline-boundary agreement by the local vessel radius from a distance transform, so that calibres are balanced. Both are the closest topology- and distance-aware competitors to our loss, which is why we benchmark against them throughout. Our loss uses warm-up $T_w=30$ epochs, weight $\lambda_{\max}=0.20$, equal term weights $\lambda_{\mathrm{rem}}=\lambda_{\mathrm{cre}}=1$, and a distance clip of $\delta=20$~mm. To isolate the surface term from extra training, we continue the Dice+CE baseline for the same $150$ epochs, serving as a matched-epoch control.
 
\subsubsection{Evaluation.}
We report three metric families: Dice coefficient for volume overlap; clDice and absolute error of the Betti number $\beta_0$ for topology; and precision, recall, and $F_1$ from our surface metric. We test each fine-tuned loss (clDice, cbDice, and ours) against the Dice+CE baseline, on both Dice and surface $F_1$, using a paired Wilcoxon signed-rank test with Holm correction across comparisons.
 
\section{Results and Discussion}
\label{sec:results}
 
\subsection{What Dice Cannot See}
\label{sec:results-dice}
We substantiate the three limitations of Sec.~\ref{sec:intro} by applying our surface metric (Sec.~\ref{sec:metric}) to the strong Dice+CE baseline over the $200$ held-out test cases, never used for training or hyperparameter selection. At a mean Dice of $80.99$, the baseline still omits over a third of the thinnest branches and fragments the tree. First, the error is \emph{directional}, which symmetric Dice masks: the baseline produces a median of $14{,}972$ spurious surface points (FP) against $22{,}124$ missed ones (FN)---a median per-case FN/FP ratio of $1.45$, with FN exceeding FP in $72.5\%$ of cases---so it omits far more vessel than it hallucinates, yet Dice collapses both into one value. Second, the error \emph{concentrates in distal branches that carry little volume} (Table~\ref{tab:radius}): thin branches ($r<0.8$~mm) are missed at $39.5\%$, nearly an order of magnitude above thick proximal segments ($r\ge1.5$~mm) at $4.5\%$, yet hold only $1.2\%$ of vessel volume against $74.9\%$. A volume-weighted overlap is thus dominated by the proximal regions the model already gets right, blind to distal failure. Finally, these misses \emph{fragment topology}, deviating from the reference by $6.14\pm3.6$ connected components ($\beta_0$-err), a defect orthogonal to overlap. These are intrinsic properties of Dice: being volume-weighted and symmetric, it underweights distal boundaries and conflates error types, so any decoupled metric would expose the same gap.

 
\begin{table}[h]
\centering
\caption{Surface error by vessel calibre on the Dice+CE baseline ($n=200$). The missed
(false-negative) rate rises sharply toward thin distal branches, yet those branches
carry a negligible share of the volume that drives Dice. Radius from the reference
distance transform; per-voxel volume share assigned by nearest medial-axis radius.}
\label{tab:radius}
\setlength{\tabcolsep}{6pt}
\begin{tabular}{lcc}
\toprule
Vessel calibre & Missed (FN) rate $\downarrow$ & Vol.\ share\\
\midrule
Thin distal ($r<0.8$~mm)      & $39.5\%$ & $1.2\%$\\
Medium ($0.8\le r<1.5$~mm)    & $12.5\%$ & $24.0\%$\\
Thick proximal ($r\ge1.5$~mm) & $4.5\%$  & $74.9\%$\\
\bottomrule
\end{tabular}
\end{table}
 
\subsection{Main Comparison}
\label{sec:results-main}
 
\begin{table}[t]
\centering
\caption{ImageCAS comparison across backbones. Dice, clDice, $F_1$ higher is better;
$\beta_0$-err, FP (spurious), FN (missed) lower is better. The decoupled FP/FN show
\emph{which} error each method reduces. Dice/clDice/$F_1$ on a $0$--$100$ scale; bold =
best per backbone group. $\ddagger$/$\dagger$: significant over Dice+CE on Dice/$F_1$
(Wilcoxon, $p<0.05$, Holm).}
\label{tab:imagecas}
\setlength{\tabcolsep}{4pt}
\begin{tabular}{llcccccc}
\toprule
Backbone & Loss & Dice $\uparrow$ & clDice $\uparrow$ & $\beta_0$-err $\downarrow$ &
FP $\downarrow$ & FN $\downarrow$ & $F_1\uparrow$\\
\midrule
\multirow{4}{*}{nnU-Net}
  & Dice+CE                  & 80.99 & 86.23 & 6.14 & $14{,}972$ & $22{,}124$ & 79.50\\
  & clDice~\cite{cldice2021} & 81.64$^\ddagger$ & 86.78 & 5.14 & $14{,}180$ & $21{,}542$ & 80.20$^\dagger$\\
  & cbDice~\cite{cbdice2024} & 81.92$^\ddagger$ & 86.99 & 4.53 & $13{,}835$ & $21{,}089$ & 80.50$^\dagger$\\
  & \textbf{Ours}            & \textbf{82.05}$^\ddagger$ & \textbf{87.12} & \textbf{4.05} & $\mathbf{13{,}215}$ & $\mathbf{15{,}798}$ & \textbf{82.60}$^\dagger$\\
\midrule
\multirow{4}{*}{SwinUNETR}
  & Dice+CE                  & 79.23 & 83.28 & 20.44 & $19{,}717$ & $24{,}215$ & 77.30\\
  & clDice~\cite{cldice2021} & 80.33$^\ddagger$ & 85.14 & 14.42 & $19{,}354$ & $23{,}985$ & 78.50$^\dagger$\\
  & cbDice~\cite{cbdice2024} & 80.57$^\ddagger$ & 85.94 & 13.12 & $19{,}068$ & $23{,}657$ & 78.70$^\dagger$\\
  & \textbf{Ours}            & \textbf{80.93}$^\ddagger$ & \textbf{86.23} & \textbf{11.25} & $\mathbf{17{,}540}$ & $\mathbf{19{,}853}$ & \textbf{79.70}$^\dagger$\\
\midrule
\multirow{4}{*}{NexToU}
  & Dice+CE                  & 80.80 & 86.47 & 6.39 & $14{,}818$ & $20{,}215$ & 79.60\\
  & clDice~\cite{cldice2021} & 81.50$^\ddagger$ & 86.70 & 5.29 & $14{,}385$ & $19{,}231$ & 79.90$^\dagger$\\
  & cbDice~\cite{cbdice2024} & 81.63$^\ddagger$ & 86.94 & 5.02 & $14{,}022$ & $18{,}963$ & 80.10$^\dagger$\\
  & \textbf{Ours}            & \textbf{81.94}$^\ddagger$ & \textbf{87.52} & \textbf{4.35} & $\mathbf{13{,}098}$ & $\mathbf{16{,}224}$ & \textbf{82.10}$^\dagger$\\
\bottomrule
\end{tabular}
\end{table}
 
\begin{table}[t]
\centering
\caption{ASOCA second-dataset replication, strongest baseline vs.\ our method per
backbone. Columns as in Table~\ref{tab:imagecas}. With $n=8$ test cases we report point
estimates only; the consistent direction corroborates the ImageCAS findings.}
\label{tab:asoca}
\setlength{\tabcolsep}{4pt}
\begin{tabular}{llcccccc}
\toprule
Backbone & Method & Dice $\uparrow$ & clDice $\uparrow$ & $\beta_0$-err $\downarrow$ &
FP $\downarrow$ & FN $\downarrow$ & $F_1\uparrow$\\
\midrule
\multirow{2}{*}{nnU-Net}
  & Best baseline & 88.14 & 87.81 & 8.75 & $4{,}270$ & $5{,}157$ & 85.40\\
  & \textbf{Ours} & \textbf{88.54} & \textbf{87.98} & \textbf{7.22} & $\mathbf{3{,}985}$ & $\mathbf{4{,}520}$ & \textbf{87.10}\\
\midrule
\multirow{2}{*}{SwinUNETR}
  & Best baseline & 85.11 & 84.77 & 15.62 & $4{,}663$ & $8{,}288$ & 81.90\\
  & \textbf{Ours} & \textbf{85.74} & \textbf{85.12} & \textbf{13.25} & $\mathbf{4{,}283}$ & $\mathbf{7{,}235}$ & \textbf{82.50}\\
\midrule
\multirow{2}{*}{NexToU}
  & Best baseline & 86.89 & 86.27 & 7.88 & $4{,}559$ & $5{,}639$ & 84.30\\
  & \textbf{Ours} & \textbf{87.25} & \textbf{86.90} & \textbf{6.54} & $\mathbf{4{,}231}$ & $\mathbf{4{,}725}$ & \textbf{84.90}\\
\bottomrule
\end{tabular}
\end{table}
 
On nnU-Net, fine-tuning with the surface loss yields only a small volume-overlap change (Table~\ref{tab:imagecas}): Dice edges up from $80.99$ (matched-epoch control) to $82.05$, statistically significant yet modest next to the surface change the metric reveals. Surface $F_1$ rises from $79.50$ to $82.60$, driven by improved recall as the creation term recovers missed distal vessel; the gain is significant on all three backbones ($p<0.05$, Holm-corrected; $p=1.2\times10^{-3}$ on nnU-Net, $n=200$). Because the baseline is a matched-epoch control trained for the same extra epochs, this gain is attributable to the surface objective rather than to longer training. The decoupled counts locate it: false negatives drop $22{,}124\!\to\!15{,}798$ ($-29\%$) while false positives also fall $14{,}972\!\to\!13{,}215$, so the loss recovers missed structure without inventing spurious mass. Notably cbDice, the closest prior distance-transform loss, leaves false negatives at $21{,}089$ against our $15{,}798$: a $>\!5{,}000$-point gap showing the creation term captures thin distal boundaries that symmetric distance maps ignore. The same directional trend is observed on ASOCA (Table~\ref{tab:asoca}),
providing supporting evidence on a second public dataset. However, given the small test set ($n=8$), these results should be interpreted as corroborative rather than as strong external validation.
 
\begin{figure}[h]
\centering
\includegraphics[width=0.83\textwidth]{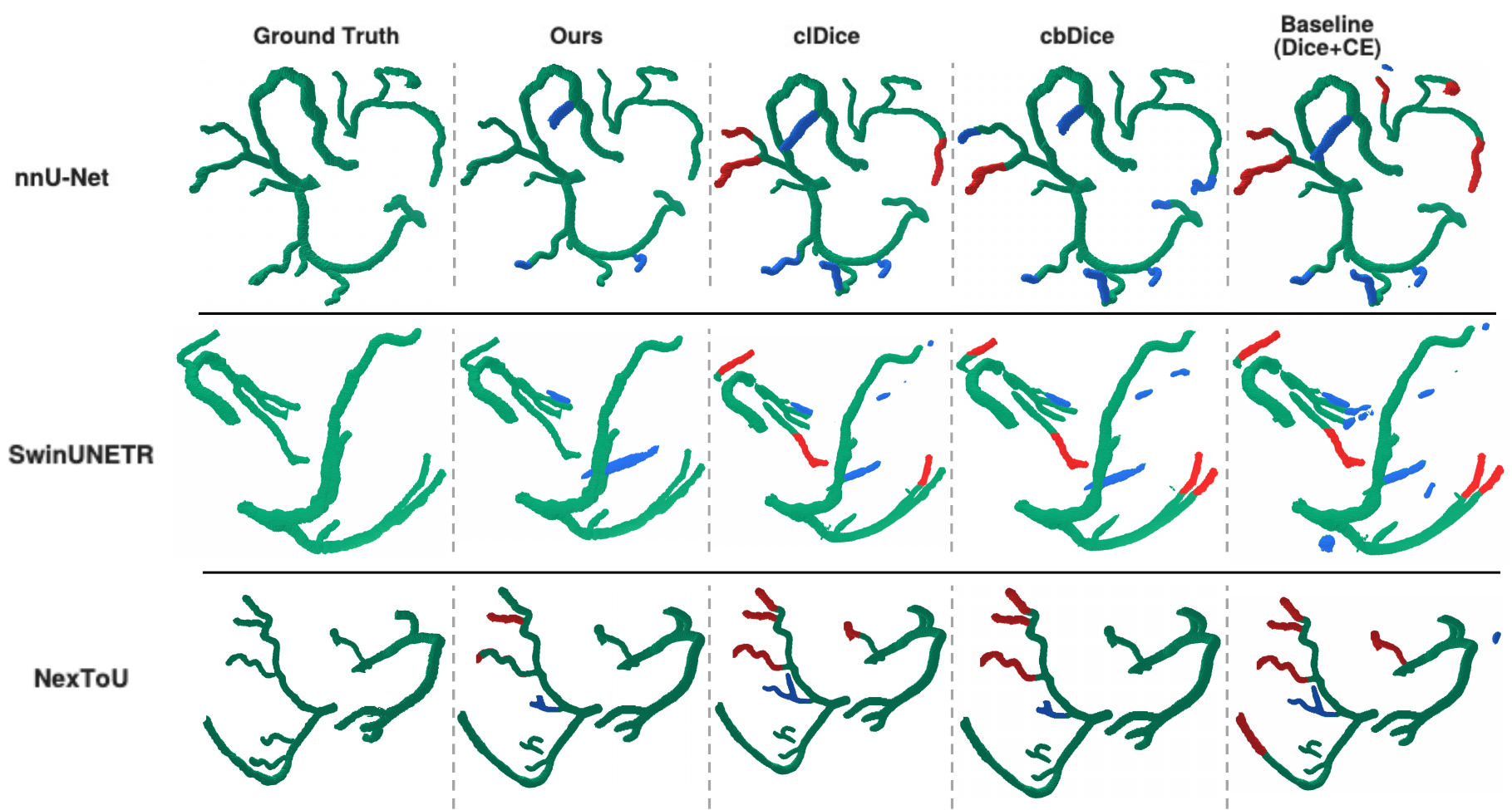}
\caption{Qualitative comparison across the three backbones (rows) and the competing
objectives (columns: reference, our method, clDice, cbDice, and the Dice+CE baseline).
Predictions are coloured against the reference: green = correctly segmented (TP), red =
missed reference vessel (FN), blue = spurious predicted vessel (FP). Our loss recovers
missed terminal vessel (less red) across all three backbones while keeping over-segmentation
(blue) rare.}
\label{fig:qualitative}
\end{figure}

Figure~\ref{fig:qualitative} makes this visible. Predictions are coloured against the reference (green: TP; red: missed FN; blue: spurious FP). Across all three backbones the overlap-trained baselines leave many thin distal branches unsegmented (large red regions); our loss recovers a substantial fraction of them, restoring tree continuity, while blue over-segmentation stays rare.
 
\subsection{Metric Validation and Sensitivity Analysis}
A value on real segmentations does not by itself establish that the metric measures what it claims. We validate it through controlled perturbations of real reference masks, where the correct response is known in advance, and probe its stability under the two free choices in its computation: the matching tolerance and the surface sampling density. These experiments characterise the metric, not model performance. Controlled perturbations need only reference masks and use the $200$ held-out test masks; the tolerance and sampling analyses need predictions and use the $160$ validation segmentations of the first fold.

\textbf{Controlled perturbations.} Degrading a reference mask (used as a perfect prediction) in known ways isolates the metric's behaviour from any model's. (i)~\emph{Identity:} prediction equal to reference gives $F_1=1$ with zero false positives and negatives on all $200$ cases. (ii)~\emph{False-negative sweep:} removing the thinnest branches (local radius below a growing threshold) drives recall down monotonically, $0.79, 0.57, 0.33, 0.18$ at $0.5, 0.8, 1.2, 1.5$~mm, while precision stays high ($0.93$--$0.98$), so the metric charges removed distal structure to false negatives. (iii)~\emph{False-positive sweep:} adding spurious mass by dilation drives precision down monotonically, $1.00, 0.73, 0.54, 0.25$ over zero to three iterations, while recall stays high, charging the added mass to false positives. The decoupled precision and recall thus respond to exactly the error mode each targets.
 
\textbf{Tolerance sensitivity.} The per-point tolerance is the local vessel radius read from the reference distance transform, not a tuned hyperparameter. To quantify the metric's dependence on it, we rescale every tolerance $r_i$ by a global factor in $\{0.5, 0.75, 1.0, 1.5, 2.0\}$ and recompute the metric for all methods on the validation fold. As expected, absolute $F_1$ values shift with the tolerance, but the ranking of the methods is preserved across the entire range, so the comparison between methods is not an artefact of the chosen radius.
 
\textbf{Surface-sampling sensitivity.} Because the metric operates on Marching Cubes surface points, we subsample both point clouds to $\{100, 75, 50, 25\}\%$ of their density and recompute $F_1$ over three random seeds. The score is stable down to aggressive subsampling, with a standard deviation below ~$F_1$ points at $50\%$ density, indicating that the metric does not depend on the exact number of surface points and that the full-resolution evaluation is not sensitive to sampling noise.
 
\subsection{Limitations and Dataset Considerations}
Our results should be read in light of known limitations of the public datasets. Recent work reports annotation inconsistencies and biases in ASOCA, particularly in secondary and tertiary branches, where they can affect measured performance more than the method itself~\cite{acebes2026}; this is especially relevant here, as our surface evaluation is sensitive to small distal discrepancies. Work on ImageCAS has likewise reported data-quality issues and refined its coronary annotations~\cite{hansen2025}. Distal-recall gains should therefore be read as improved agreement with the available annotations, not as proof of anatomically correct recovery, motivating future evaluation on datasets with independently adjudicated distal annotations.
 
\section{Conclusion}
\label{sec:discussion}
 
We showed that volume-overlap metrics like Dice introduce structural blindness on thin, tubular anatomy. Decoupling segmentation failures into independent surface-matching components via adaptive bipartite assignment revealed that modern baselines systematically omit distal vessel while retaining high volumetric scores, and our surface loss directly targets this error through its two surface-distance terms: a creation term whose gradient grows the missed branch and a removal term that suppresses spurious mass, both acting on surface points so thin distal branches are weighted like thick proximal ones. The consistent gains in surface $F_1$ and connectivity across three architectures on ImageCAS and ASOCA underscore the value of prioritizing boundary geometry over volume overlap, a paradigm that extends naturally to other multi-scale networks such as hepatic and cerebrovascular trees. A further direction is the extension to native 3D point clouds: our metric and loss already operate on surface points, here from voxel masks via Marching Cubes, but the same radius-tolerant matching could apply to point sets from mesh- or point-based models directly. This would require estimating the local radius from the point cloud itself rather than a mask distance transform, which we leave to future work.
 
\begin{credits}
\subsubsection{\ackname}
This work was supported by Azure sponsorship credits granted by
Microsoft's AI for Good Research Lab. The authors gratefully acknowledge the
RISE-MICCAI Paper-Lead Mentorship Program for fostering the research
collaboration underlying this work. In particular, we thank RISE-MICCAI for
facilitating the mentorship of Rafael Velasquez by Esther Puyol-Antón, whose
guidance was instrumental throughout this project. We also gratefully acknowledge
the RISE-MICCAI Travel Grant for supporting Rafael Velasquez's attendance and
presentation of this work at MICCAI 2026.

\subsubsection{\discintname}
The authors have no competing interests to declare that are relevant to the
content of this article.
\end{credits}
 

\end{document}